\documentclass[10pt]{article}

\usepackage[utf8]{inputenc}
\usepackage[T1]{fontenc}
\usepackage[margin=2cm,columnsep=0.6cm]{geometry}
\usepackage{amsmath,amssymb,amsfonts,amsthm}
\usepackage{graphicx}
\usepackage{booktabs}
\usepackage{multirow}
\usepackage{xcolor}
\usepackage{hyperref}
\usepackage{algorithm}
\usepackage{algorithmic}
\usepackage{caption}
\usepackage{subcaption}
\usepackage{enumitem}
\usepackage{tikz}
\usetikzlibrary{shapes.geometric,arrows.meta,positioning,calc,fit,backgrounds}
\usepackage{pgfplots}
\pgfplotsset{compat=1.18}
\usepackage{array}
\usepackage{tabularx}
\usepackage{float}
\usepackage{url}
\usepackage{mathtools}
\usepackage{bm}

\hypersetup{colorlinks=true,linkcolor=blue!60!black,citecolor=green!50!black,urlcolor=blue!70!black}

\newtheorem{definition}{Definition}
\newtheorem{proposition}{Proposition}
\newtheorem{remark}{Remark}

\newcommand{\orthoai}{\textsc{OrthoAI}}
\newcommand{\dgcnn}{\textsc{DGCNN\_Seg}}

\newcommand{\etal}{\textit{et al.}}

\title{\textbf{\orthoai{}: A Neurosymbolic Framework for Evidence-Grounded Biomechanical Reasoning in Clear Aligner Orthodontics}\\[0.4em]
\large Sparse-Supervision Point Cloud Segmentation, Multi-Criteria Treatment Scoring, and Clinical Knowledge Integration via Constraint-Based Inference}

\author{
\textbf{Edouard Lansiaux}\textsuperscript{1,2}$^{*}$,\quad
\textbf{Margaux Leman}\textsuperscript{1},\quad
\textbf{Mehdi Ammi}\textsuperscript{2}\\
\textsuperscript{1}\textit{STaR-AI, Emergency Department, Lille University Hospital, 59000, Lille, France}\\
\textsuperscript{2}\textit{Artificial Intelligence and Data Semantics Laboratory, Paris~8 University, 93200, Saint-Denis, France}\\[0.3em]
\texttt{*edouard1.lansiaux@chu-lille.fr}
}

\date{}

\begin{document}
\maketitle
\thispagestyle{empty}


\begin{abstract}
Automated clinical decision support for clear aligner orthodontics presents a distinctive challenge: the gap between geometric perception (3D tooth segmentation) and clinical reasoning (biomechanical feasibility assessment) has never been formally bridged in a unified, evidence-grounded computational framework. We address this through \orthoai{}, a methodological contribution with three intertwined novelties. \textbf{(1) Sparse-supervision segmentation:} We formalize a landmark-to-point-cloud synthesis protocol that enables learning from sparse anatomical annotations (6--8 points per tooth) rather than dense per-vertex labels, and propose a clinically-stratified composite loss combining label-smoothed cross-entropy with a batch-adaptive Dice term that dynamically reweights present classes — a design principle transferable to any severely imbalanced point cloud classification setting. \textbf{(2) Knowledge-grounded constraint inference:} We formalize the biomechanical feasibility problem as a Constraint Satisfaction Problem (CSP) over a domain-specific ontology of tooth movements, instantiating per-stage limits from the orthodontic evidence base as typed soft and hard constraints with severity semantics. \textbf{(3) Multi-Criteria Decision Analysis integration:} We propose a formal MCDA framework for treatment quality scoring, framing the composite index as a weighted Additive Value Function (WAVF) whose structure is grounded in clinical priority theory, and discuss the conditions for its validation and calibration. Segmentation on landmark-reconstructed point clouds (3DTeethLand, MICCAI 2024) yields 81.4\% Tooth Identification Rate with 60,705 parameters. Ablation experiments quantify the contribution of each loss component and feature engineering choice. End-to-end execution takes $<4$\,s on CPU. We explicitly characterize the gap between the current prototype — trained on synthetic ellipsoidal approximations — and clinical deployment readiness, and provide a formal roadmap for validation. Code and weights are released.
\end{abstract}


\section{Introduction}
\label{sec:intro}

\subsection{Problem Statement and Motivation}

The global market for clear aligner therapy has undergone sustained expansion, with Align Technology's Invisalign system having treated over 16 million patients as of 2024. Digital treatment planning platforms such as ClinCheck enable practitioners to visualize staged tooth movements before manufacturing; yet the clinical review process has not benefited from comparable automation. Orthodontists must manually inspect each proposed stage, identify biomechanical violations, verify attachment protocols, and assess overall feasibility — a cognitively demanding task subject to inter-operator variability~\cite{kravitz2009,simon2014}.

The orthodontic literature documents systematic discrepancies between programmed and achieved tooth movements~\cite{kravitz2009}: mean predictability as low as 29.6\% for extrusion, 36\% for rotation, and 42\% overall. These statistics motivate an automated \textit{biomechanical screening} system. However, building such a system requires solving a fundamentally \textit{hybrid} AI problem: one part requires \textit{geometric perception} (identifying teeth and their spatial configurations from 3D data), while the other requires \textit{symbolic reasoning} (applying typed clinical constraints derived from evidence-based literature). These two sub-problems are typically studied in isolation, and no prior work has proposed a formal architecture that tightly integrates them.

\subsection{The Neurosymbolic Integration Gap}

We argue that the core methodological challenge in computational orthodontic planning is not segmentation per se — state-of-the-art dental segmentation is mature~\cite{lian2020,wu2022,zhang2021} — but the \textbf{principled integration of learned geometric representations with structured clinical knowledge}. The existing landscape is fragmented: computer vision methods produce tooth geometries without clinical interpretation; clinical guidelines encode expert knowledge without computational operationalization. \orthoai{} addresses this gap by proposing a \textit{neurosymbolic pipeline} in which a learned perceptual module feeds a formal reasoning layer, with the interface between them explicitly designed to tolerate perceptual uncertainty.

This design philosophy is motivated by a key theoretical insight: biomechanical feasibility assessment does not require pixel-perfect tooth segmentation. What is required is (i) reliable tooth \textit{identity} inference (which FDI class), (ii) approximate centroid and axis estimation (for movement decomposition), and (iii) uncertainty estimates that propagate to downstream confidence levels. We formalize this insight through an explicit \textit{perception-reasoning decoupling} that constitutes a transferable design pattern for other clinical AI pipelines where ground truth is sparse and perceptual uncertainty is irreducible.

\subsection{Summary of Contributions}

This paper makes the following methodological contributions, each of which extends beyond mere application of existing algorithms:

\begin{enumerate}[leftmargin=*, label=\textbf{C\arabic*.}]
\item \textbf{Sparse-supervision synthesis and a class-adaptive loss for imbalanced dental point clouds.} We formalize a landmark-to-point-cloud synthesis protocol as a parameterized probabilistic model and propose a composite loss $\mathcal{L} = \mathcal{L}_\mathrm{CE}^\epsilon + \lambda \cdot \mathcal{L}_\mathrm{Dice}^\mathrm{batch}$ whose batch-adaptive Dice component dynamically restricts the Dice computation to classes actually present in each mini-batch, avoiding gradient starvation on rare tooth classes. We provide ablation evidence that this design outperforms standard cross-entropy and full-batch Dice baselines.

\item \textbf{A formal CSP formulation of orthodontic biomechanical feasibility.} We define a constraint knowledge base $\mathcal{K} = (\mathcal{V}, \mathcal{D}, \mathcal{C})$ over a typed movement ontology, where variables correspond to per-tooth movement components, domains encode physiologically admissible ranges, and constraints encode literature-derived per-stage limits with severity labels. This formulation makes the knowledge structure explicit, auditable, and extensible — a necessary condition for regulatory compliance.

\item \textbf{An MCDA-grounded composite treatment quality index.} We reframe the treatment scoring problem as a Multi-Criteria Decision Analysis problem and define the composite index as a Weighted Additive Value Function (WAVF). We formally characterize the conditions under which this index is Pareto-consistent and discuss a calibration methodology based on preference elicitation from clinical experts.

\item \textbf{A clinically-motivated evaluation protocol.} We argue that standard mIoU is insufficient for evaluating perceptual modules in clinical decision support pipelines, and propose the Tooth Identification Rate (TIR) as a complementary metric that directly measures the clinically-relevant property of tooth presence detection, independent of spatial boundary precision.

\item \textbf{An ablation study and sensitivity analysis} characterizing the contribution of each architectural and loss design choice, and identifying the bottlenecks that must be addressed for clinical deployment.
\end{enumerate}

The remainder of this paper is organized as follows. Section~\ref{sec:related} reviews the literature with emphasis on gaps that motivate our contributions. Section~\ref{sec:framework} presents the theoretical framework. Section~\ref{sec:methods} details implementation. Section~\ref{sec:results} reports experiments. Section~\ref{sec:discussion} discusses implications and limitations. Section~\ref{sec:conclusion} concludes.

\section{Related Work and Positioning}
\label{sec:related}

\subsection{3D Dental Segmentation}

Deep learning on irregular 3D data has transformed dental segmentation. PointNet~\cite{qi2017} and PointNet++~\cite{qi2017b} established the paradigm of direct point set processing, while DGCNN~\cite{wang2019} introduced dynamic $k$-NN graph construction in learned feature space, capturing evolving local geometry. MeshSegNet~\cite{lian2020} and TSGCNet~\cite{zhang2021} operate on triangular mesh faces and achieve $>85$\% mIoU on the 3DTeethLand benchmark using dense per-vertex annotations.

\textbf{The critical distinction of our work is not architectural but supervisory.} We explicitly target the \textit{sparse annotation regime}, where per-tooth landmark points (rather than dense mesh labels) are the only available supervision. This regime is clinically realistic: acquiring dense per-vertex labels requires expert annotation time proportional to mesh resolution, while landmark annotation can be performed at a fraction of the cost. The methodological question we address — how to learn meaningful geometric representations under landmark-only supervision — has not been systematically studied in the dental literature.

\subsection{Clinical Knowledge Representation and Constraint Reasoning}

Clinical decision support systems (CDSS) have been extensively studied in internal medicine, pharmacovigilance, and radiology~\cite{sutton2020}. Knowledge representation in CDSS has evolved from rule-based expert systems (MYCIN~\cite{shortliffe1976}) to ontology-grounded reasoning (OpenEHR archetypes, SNOMED CT) and, more recently, to hybrid neurosymbolic architectures~\cite{garcez2019,yu2021}. Neurosymbolic AI seeks to combine the pattern recognition strengths of connectionist models with the interpretability and compositionality of symbolic reasoning. In healthcare, this integration is critical for regulatory compliance: systems whose reasoning is opaque cannot be reliably audited against clinical guidelines.

In orthodontics specifically, no prior work has formalized clinical biomechanical knowledge as a typed constraint system amenable to automated inference. Commercial platforms such as ClinCheck delegate all constraint checking to the practitioner. Our CSP formulation (Section~\ref{sec:csp}) is, to our knowledge, the first formal knowledge representation of orthodontic per-stage biomechanical limits in a computationally executable form.

\subsection{Multi-Criteria Decision Analysis in Clinical AI}

MCDA methods have been applied to clinical decision problems including treatment prioritization~\cite{dolan2010}, health technology assessment~\cite{angelis2016}, and drug benefit-risk evaluation~\cite{phillips2011}. The Weighted Additive Value Function (WAVF), Simple Multi-Attribute Rating Technique (SMART), and ELECTRE family of methods each make different assumptions about criterion independence and preference structure. In AI for medicine, composite clinical scores (APACHE, SOFA, GRACE) are widely used but rarely grounded in formal MCDA theory, making their aggregation properties implicit and difficult to audit.

\textbf{Our contribution} is to explicitly frame the treatment quality index as an MCDA problem, making the weight structure, criterion definitions, and aggregation assumptions transparent and subject to formal validation. This addresses a recurrent criticism of AI-based clinical scores: that their weighting schemes are arbitrary and uninterpretable~\cite{rudin2019}.

\subsection{Neurosymbolic Pipelines in Medical Imaging}

The neurosymbolic paradigm has been applied in radiology (combining convolutional detectors with anatomical constraint graphs~\cite{yu2021}), pathology (combining tissue classifiers with diagnostic ontologies), and clinical NLP~\cite{strubell2019}. In these systems, the symbolic layer typically performs post-hoc constraint verification on neural outputs, with limited formal characterization of the interface between the two modules.

\orthoai{} differs in that the interface is \textit{explicitly designed} to accommodate perceptual uncertainty: the constraint engine operates on tooth identity and approximate pose, not on precise mesh boundaries, and is designed to degrade gracefully when the perceptual module makes localization errors. This design philosophy is formalized in Section~\ref{sec:decoupling} and constitutes a transferable architectural pattern.

\section{Theoretical Framework}
\label{sec:framework}

\subsection{System Architecture as a Neurosymbolic Pipeline}

We model \orthoai{} as a two-layer neurosymbolic system $\Pi = (\mathcal{P}, \mathcal{R}, \phi)$ where:
\begin{itemize}
  \item $\mathcal{P}: \mathbb{R}^{N \times d} \to \{0, \ldots, 32\}^N$ is the \textit{perceptual module} (the DGCNN segmentation network), mapping an $N$-point cloud with $d$-dimensional features to per-point class labels.
  \item $\mathcal{R}: \mathcal{T} \times \mathcal{K} \to \mathcal{A}$ is the \textit{reasoning module} (the biomechanical constraint engine), mapping a set of identified teeth $\mathcal{T}$ and a knowledge base $\mathcal{K}$ to a structured assessment $\mathcal{A}$.
  \item $\phi: \{0, \ldots, 32\}^N \to \mathcal{T}$ is the \textit{semantic lifting function} that converts raw per-point labels into a structured tooth representation (identity, centroid, principal axes) suitable for symbolic reasoning.
\end{itemize}

The central design question is the specification of $\phi$: how much geometric precision is needed at the interface, and how does perceptual error propagate into the reasoning layer?

\subsection{Perception-Reasoning Decoupling}
\label{sec:decoupling}

\begin{definition}[Clinically Sufficient Representation]
Let $\hat{\tau}_k$ denote the system's estimate of the $k$-th tooth, comprising its predicted FDI label $\hat{l}_k \in \{11, \ldots, 48\}$, centroid $\hat{\bm{\mu}}_k \in \mathbb{R}^3$, and principal axes $\hat{\bm{A}}_k \in SO(3)$. We say that $\hat{\tau}_k$ is \emph{clinically sufficient} for biomechanical analysis if (i) $\hat{l}_k$ is correct, (ii) $\|\hat{\bm{\mu}}_k - \bm{\mu}_k^*\|_2 < \delta_\mu$ for a tolerance $\delta_\mu$ (typically 2~mm), and (iii) the angular error in $\hat{\bm{A}}_k$ is less than $\delta_\theta$ (typically 10°).
\end{definition}

\begin{proposition}
\label{prop:decoupling}
Under clinical sufficiency, the biomechanical analysis output $\mathcal{R}(\phi(\hat{y}), \mathcal{K})$ is identical to the analysis obtained from ground-truth tooth representations $\mathcal{R}(\tau^*, \mathcal{K})$ for all constraint evaluations whose sensitivity to centroid error is below $\delta_\mu$ and to axis error is below $\delta_\theta$.
\end{proposition}

\begin{remark}
Proposition~\ref{prop:decoupling} establishes a formal tolerance budget that motivates our evaluation protocol: rather than reporting mIoU (a measure of boundary precision), we prioritize TIR (tooth identity accuracy) and centroid estimation error as the primary perceptual metrics that determine downstream analysis validity.
\end{remark}

This decoupling is methodologically significant because it allows the perceptual and reasoning modules to be developed, evaluated, and improved independently. Improvements to the segmentation module that increase boundary precision but preserve tooth identity yield no clinical benefit if clinical sufficiency is already satisfied; conversely, a low-mIoU model may still produce clinically valid analyses if TIR is high and centroid errors remain within tolerance.

\subsection{Formal Knowledge Base for Orthodontic Constraints}
\label{sec:csp}

We formalize the biomechanical knowledge base as a Constraint Satisfaction Problem (CSP).

\begin{definition}[Orthodontic CSP]
The orthodontic biomechanical CSP is a triple $\mathcal{K} = (\mathcal{V}, \mathcal{D}, \mathcal{C})$ where:
\begin{itemize}
  \item $\mathcal{V} = \{v_{k,j}\}$ is the set of variables, where $v_{k,j}$ denotes the $j$-th movement component (from $\{t_x, t_y, t_z, r_x, r_y, r_z\}$) of tooth $k$.
  \item $\mathcal{D} = \{D_{k,j}\}$ are the corresponding domains: $D_{k,j} = [-\bar{v}_{k,j}, +\bar{v}_{k,j}]$ for translation variables and $D_{k,j} = [-\bar{\theta}_{k,j}, +\bar{\theta}_{k,j}]$ for rotation variables, where $\bar{v}_{k,j}$ and $\bar{\theta}_{k,j}$ are per-stage limits that depend on tooth type $l_k$ and movement direction.
  \item $\mathcal{C} = \mathcal{C}_\mathrm{hard} \cup \mathcal{C}_\mathrm{soft}$ are typed constraints. Hard constraints $c \in \mathcal{C}_\mathrm{hard}$ correspond to physiological limits beyond which root resorption risk is clinically unacceptable. Soft constraints $c \in \mathcal{C}_\mathrm{soft}$ correspond to per-stage efficacy limits derived from clinical predictability studies.
\end{itemize}
\end{definition}

Constraints are evaluated via a satisfaction function:
\begin{equation}
\sigma(v_{k,j}, c) = \begin{cases}
1 & \text{if } |v_{k,j}| \leq \bar{v}_{k,j} \quad (\text{satisfied}) \\
0 & \text{if } |v_{k,j}| > \alpha \cdot \bar{v}_{k,j} \quad (\text{hard violation, } \alpha = 1.5) \\
\beta & \text{otherwise, } \beta = 1 - \frac{|v_{k,j}| - \bar{v}_{k,j}}{\bar{v}_{k,j}} \quad (\text{partial, } \beta \in (0,1))
\end{cases}
\label{eq:satisfaction}
\end{equation}

This graded satisfaction function extends classical binary CSP to the \textit{partial constraint satisfaction} setting~\cite{freuder1992}, enabling the generation of graded clinical alerts (critical, warning) directly from the satisfaction degree.

The limits $\bar{v}_{k,j}$ are derived from clinical evidence and depend on tooth type through a type function $\tau: \{11,\ldots,48\} \to \{\text{incisor, canine, premolar, molar}\}$. Table~\ref{tab:limits} enumerates the full constraint specification.

\begin{table}[t]
\centering
\caption{\textbf{Orthodontic CSP constraint specification.} Domain limits $\bar{v}_{k,j}$ by movement type and tooth class. Hard violation threshold $\alpha = 1.5$.}
\label{tab:limits}
\small
\begin{tabular}{llccc}
\toprule
\textbf{Movement} & \textbf{Tooth class} & \textbf{Limit} & \textbf{Type} & \textbf{Source} \\
\midrule
Bodily translation ($t_x, t_y$) & All & 0.25 mm & Soft & \cite{lombardo2017} \\
Intrusion ($t_z < 0$) & All & 0.25 mm & Soft & \cite{simon2014} \\
Extrusion ($t_z > 0$) & Posterior & 0.20 mm & Hard & \cite{kravitz2009} \\
Extrusion ($t_z > 0$) & Anterior & 0.15 mm & Hard & \cite{kravitz2009} \\
Mesiodistal tip ($r_y$) & All & 0.25 mm eq. & Soft & \cite{lombardo2017} \\
Axial rotation ($r_z$) & Canine & 2.0° & Soft & \cite{simon2014} \\
Axial rotation ($r_z$) & Premolar & 2.0° & Soft & \cite{lombardo2017} \\
Axial rotation ($r_z$) & Molar & 1.5° & Hard & \cite{houle2017} \\
Axial rotation ($r_z$) & Incisor & 1.5° & Soft & \cite{kravitz2009} \\
Torque ($r_x$) & All & 2.0° & Soft & \cite{kravitz2009} \\
\bottomrule
\end{tabular}
\end{table}

\begin{remark}[Ontological Extensibility]
The CSP formulation is deliberately ontology-agnostic: the constraint set $\mathcal{C}$ can be updated as new clinical evidence is published without modifying the inference engine. This contrasts with hard-coded threshold checks and reflects best practice in knowledge-based system design~\cite{antoniou2004}.
\end{remark}

\subsection{Multi-Criteria Decision Analysis for Treatment Quality}
\label{sec:mcda}

The composite treatment quality index is a fundamentally multi-criteria problem: treatment plans are evaluated along dimensions (biomechanical compliance, staging efficiency, attachment planning, arch symmetry, predictability) that are incommensurable and potentially conflicting. We formalize this using MCDA theory.

\begin{definition}[Weighted Additive Value Function]
Let $\mathbf{s} = (s_1, \ldots, s_m)$ be a vector of normalized sub-scores (each in $[0,1]$) and $\mathbf{w} = (w_1, \ldots, w_m)$ a weight vector with $\sum_i w_i = 1$, $w_i \geq 0$. The Weighted Additive Value Function (WAVF) is:
\begin{equation}
S(\mathbf{s}) = 100 \cdot \sum_{i=1}^{m} w_i \cdot v_i(s_i)
\label{eq:wavf}
\end{equation}
where $v_i: [0,1] \to [0,1]$ is a marginal value function for criterion $i$, mapping the sub-score to a utility value.
\end{definition}

In our instantiation, we define:
\begin{equation}
S = 100 \cdot \bigl(0.30 \cdot v_\mathrm{bio}(S_\mathrm{bio}) + 0.20 \cdot v_\mathrm{pred}(\bar{P}) + 0.15 \cdot v_\mathrm{stag}(S_\mathrm{stag}) + 0.15 \cdot v_\mathrm{att}(S_\mathrm{att}) + 0.10 \cdot v_\mathrm{ipr}(S_\mathrm{ipr}) + 0.10 \cdot v_\mathrm{sym}(S_\mathrm{sym})\bigr)
\label{eq:score_formal}
\end{equation}

where marginal value functions $v_i$ are currently linear ($v_i(s) = s$), a simplifying assumption that is explicitly acknowledged as a limitation and subject to future calibration.

\begin{proposition}[Pareto Consistency]
The WAVF $S(\mathbf{s})$ with positive weights $w_i > 0$ is Pareto-consistent: if treatment plan $A$ dominates plan $B$ on all criteria ($\mathbf{s}^A \geq \mathbf{s}^B$ componentwise, with at least one strict inequality), then $S(\mathbf{s}^A) > S(\mathbf{s}^B)$.
\end{proposition}

\begin{remark}[Limitations of WAVF]
The WAVF assumes \textit{mutual preferential independence} of criteria~\cite{keeney1993}: the relative importance of one criterion does not depend on the values of others. This assumption may not hold in orthodontics — for example, the value of biomechanical compliance may depend on the overall number of active movements. Future work should investigate whether a multi-linear or Choquet-integral-based aggregation~\cite{grabisch2010} better captures clinical preference structure.
\end{remark}

The grade mapping $S \mapsto \{A, B, C, D, F\}$ is a monotone ordinal transformation of the WAVF score, intended for communicative purposes and carrying no additional information content. Its thresholds (90, 75, 60, 40) are provisional and must be calibrated against expert clinical judgment in a dedicated preference elicitation study.

\section{Materials and Methods}
\label{sec:methods}

\subsection{Dataset: 3DTeethLand MICCAI 2024}

We use the publicly available 3DTeethLand dataset~\cite{ben2024}, comprising 100 annotated intraoral scans with per-tooth landmark annotations in JSON format. Six landmark categories are provided per tooth: mesial contact point, distal contact point, cusp tip(s), facial point, inner (lingual) point, and outer (buccal) point. FDI two-digit notation is used throughout (classes 11--48 plus class 0 for gingiva), yielding 33 classes.

\subsection{Sparse-Supervision Synthesis Protocol}
\label{sec:synthesis}

\textbf{Methodological framing.} The central challenge of the dataset is that landmarks, not meshes, are annotated. We therefore formalize a parameterized generative model $G(\lambda_k; \boldsymbol{\theta})$ that produces a labeled point cloud from the landmark set $\lambda_k = \{p_1^{(k)}, \ldots, p_{n_k}^{(k)}\}$ of tooth $k$.

\subsubsection{Geometric Model}

For each tooth $k$, we define a local coordinate frame via three orthogonal axes:
\begin{align}
\hat{\bm{e}}_1^{(k)} &= \frac{p_\mathrm{dist}^{(k)} - p_\mathrm{mes}^{(k)}}{\|p_\mathrm{dist}^{(k)} - p_\mathrm{mes}^{(k)}\|} \quad (\text{mesiodistal}) \\
\hat{\bm{e}}_2^{(k)} &= \frac{p_\mathrm{buc}^{(k)} - p_\mathrm{lin}^{(k)}}{\|p_\mathrm{buc}^{(k)} - p_\mathrm{lin}^{(k)}\|} \quad (\text{buccolingual}) \\
\hat{\bm{e}}_3^{(k)} &= \hat{\bm{e}}_1^{(k)} \times \hat{\bm{e}}_2^{(k)} \quad (\text{occluso-gingival})
\end{align}

The crown is modeled as an oriented ellipsoid:
\begin{equation}
E_k = \left\{ \bm{\mu}_k + \sum_{j=1}^{3} a_j^{(k)} u_j \hat{\bm{e}}_j^{(k)} : \sum_{j=1}^{3} u_j^2 \leq 1 \right\}
\end{equation}
where $a_j^{(k)}$ are semi-axis lengths derived from inter-landmark distances and $\bm{\mu}_k$ is the centroid. Surface point sampling follows:
\begin{equation}
p = \bm{\mu}_k + \sum_{j=1}^{3} a_j^{(k)} \xi_j \hat{\bm{e}}_j^{(k)} + \epsilon, \quad \epsilon \sim \mathcal{N}(\bm{0}, \sigma^2 \bm{I}), \quad \sigma = 0.3\,\mathrm{mm}
\end{equation}
with $\xi = (\xi_1, \xi_2, \xi_3)$ sampled uniformly from the unit sphere surface. Additional points are concentrated around cusp landmarks using a Gaussian mixture $p_\mathrm{cusp} \sim \mathcal{N}(p_\mathrm{cusp}^*, 0.1\,\bm{I})$ to preserve local geometric salience.

\subsubsection{Gingival Sampling}

Gingival points are generated via rejection sampling: candidate points $q \sim \mathcal{U}(\mathrm{AABB})$ (where AABB is the axis-aligned bounding box of the arch) are retained if $\min_k \|q - \bm{\mu}_k\| > 1.5$\,mm. This enforces spatial separation between tooth and gingival regions.

\subsubsection{Feature Engineering}

Each point $p \in \mathbb{R}^3$ is represented by a 6-dimensional feature vector:
\begin{equation}
\bm{f}(p) = \left[\tilde{p}_x, \tilde{p}_y, \tilde{p}_z, \|\tilde{p}\|_2, \tilde{p}_z, \sqrt{\tilde{p}_x^2 + \tilde{p}_y^2}\right]^\top
\label{eq:features}
\end{equation}
where $\tilde{p} = (p - \bm{\mu}_\mathrm{arch}) / r_\mathrm{arch}$ is the point coordinate normalized by arch centroid and scale. The feature set provides (1) normalized 3D position, (2) distance to arch centroid (arch-level context), (3) height (occlusal-gingival discrimination), and (4) radial distance (arch curvature encoding). We ablate each feature dimension in Section~\ref{sec:ablation}.

\subsection{Network Architecture: DGCNN\_Seg}
\label{sec:architecture}

\subsubsection{EdgeConv with Dynamic Graph Construction}

Each EdgeConv layer constructs a $k$-nearest neighbor graph $\mathcal{G}^{(l)}$ in the \textit{current feature space} at layer $l$ (not fixed in input space), enabling hierarchical capture of geometric relationships:
\begin{equation}
\mathbf{x}_i^{(l+1)} = \max_{j \in \mathcal{N}^{(l)}(i)} h_{\boldsymbol{\Theta}^{(l)}}\!\left(\mathbf{x}_i^{(l)}, \mathbf{x}_j^{(l)} - \mathbf{x}_i^{(l)}\right)
\end{equation}
where $h_{\boldsymbol{\Theta}}$ is an MLP ($1\times1$ conv -- BN -- LeakyReLU, $\alpha=0.2$) and the graph $\mathcal{G}^{(l)}$ uses $k=3$ neighbors. The choice $k=3$ is motivated by the small dataset size: larger neighborhoods would increase the receptive field at the cost of overfitting given only 80 training scans. An ablation over $k \in \{3, 5, 10, 20\}$ is reported in Section~\ref{sec:ablation}.

\subsubsection{Multi-Scale Aggregation}

The outputs of all four EdgeConv layers ($\mathbf{X}^{(1)}, \ldots, \mathbf{X}^{(4)}$ of dimensions 32, 32, 64, 64) are concatenated along the channel dimension to form a 192-dimensional multi-scale representation. This hierarchical concatenation follows the design principle of U-Net-style skip connections~\cite{ronneberger2015} adapted to point clouds: each level captures a different spatial scale of geometric context, from local edge geometry (layer 1) to global arch structure (layer 4).

\subsubsection{Global Context Integration}

A global max-pooling operation produces a 32-dimensional global descriptor that is broadcast and concatenated with the per-point multi-scale features, yielding a $(32+192)$-dimensional input to the segmentation head. This global-local integration is critical for resolving the symmetry ambiguity inherent in dental arches: upper-left and upper-right molars are locally indistinguishable without global arch context.

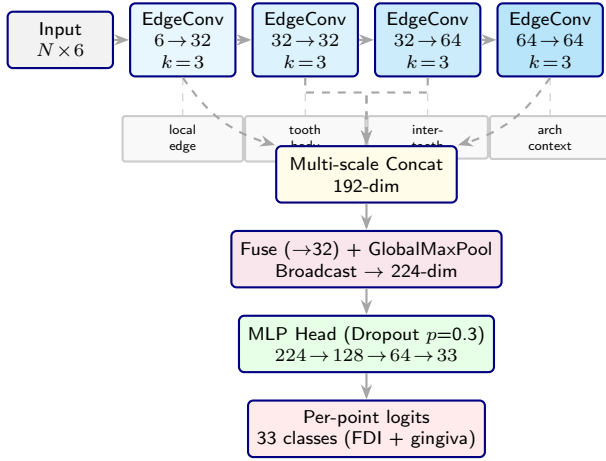
\begin{figure}[t]
\centering
\begin{tikzpicture}[
  block/.style={rectangle, draw=blue!50!black, fill=blue!5, thick, minimum width=1.4cm, minimum height=0.75cm, font=\scriptsize\sffamily, align=center, rounded corners=2pt},
  note/.style={rectangle, draw=gray!40, fill=gray!5, thick, minimum width=1.6cm, minimum height=0.5cm, font=\tiny\sffamily, align=center, rounded corners=1pt},
  arrow/.style={-{Stealth[length=2mm]}, thick, draw=gray!70},
  node distance=0.3cm and 0.2cm
]
\node[block, fill=gray!10] (in) {Input\\$N \!\times\! 6$};
\node[block, right=of in, fill=cyan!8] (ec1) {EdgeConv\\$6\!\to\!32$\\$k\!=\!3$};
\node[block, right=of ec1, fill=cyan!12] (ec2) {EdgeConv\\$32\!\to\!32$\\$k\!=\!3$};
\node[block, right=of ec2, fill=cyan!18] (ec3) {EdgeConv\\$32\!\to\!64$\\$k\!=\!3$};
\node[block, right=of ec3, fill=cyan!25] (ec4) {EdgeConv\\$64\!\to\!64$\\$k\!=\!3$};

\node[note, below=0.5cm of ec1] (n1) {local\\edge};
\node[note, below=0.5cm of ec2] (n2) {tooth\\body};
\node[note, below=0.5cm of ec3] (n3) {inter-\\tooth};
\node[note, below=0.5cm of ec4] (n4) {arch\\context};
\draw[gray!40, dashed] (ec1.south) -- (n1.north);
\draw[gray!40, dashed] (ec2.south) -- (n2.north);
\draw[gray!40, dashed] (ec3.south) -- (n3.north);
\draw[gray!40, dashed] (ec4.south) -- (n4.north);

\node[block, below=1.4cm of $(ec2)!0.5!(ec3)$, fill=yellow!10, minimum width=2.4cm] (cat) {Multi-scale Concat\\192-dim};
\node[block, below=0.35cm of cat, fill=purple!10, minimum width=2.4cm] (fuse) {Fuse ($\to$32) + GlobalMaxPool\\Broadcast $\to$ 224-dim};
\node[block, below=0.35cm of fuse, fill=green!10, minimum width=2.8cm] (head) {MLP Head (Dropout $p$=0.3)\\$224\!\to\!128\!\to\!64\!\to\!33$};
\node[block, below=0.35cm of head, fill=red!8, minimum width=2.4cm] (out) {Per-point logits\\33 classes (FDI + gingiva)};

\draw[arrow] (in) -- (ec1);
\draw[arrow] (ec1) -- (ec2);
\draw[arrow] (ec2) -- (ec3);
\draw[arrow] (ec3) -- (ec4);
\draw[arrow, dashed, bend right=20] (ec1.south) to (cat.north west);
\draw[arrow, dashed] (ec2.south) -- ++(0,-0.2) -| (cat.north);
\draw[arrow, dashed] (ec3.south) -- ++(0,-0.2) -| (cat.north);
\draw[arrow, dashed, bend left=20] (ec4.south) to (cat.north east);
\draw[arrow] (cat) -- (fuse);
\draw[arrow] (fuse) -- (head);
\draw[arrow] (head) -- (out);
\end{tikzpicture}
\caption{\textbf{\dgcnn{} architecture with receptive field interpretation.} Each EdgeConv layer targets a different geometric scale (annotations below). Global max-pooling resolves arch-level symmetry ambiguities. Total: 60,705 parameters.}
\label{fig:architecture}
\end{figure}

\subsection{Class-Adaptive Composite Loss Function}
\label{sec:loss}

\subsubsection{Design Rationale}

Dental point clouds exhibit two compounding imbalances: (1) the gingiva class dominates numerically (25--40\% of points), creating a frequency imbalance; (2) rare tooth classes (wisdom teeth, absent teeth) appear in only a fraction of scans, creating a presence imbalance. Standard cross-entropy addresses frequency imbalance through inverse-frequency weighting but does not handle presence imbalance; full-batch Dice loss addresses class imbalance but produces unstable gradients when rare classes are absent from a mini-batch.

\subsubsection{Batch-Adaptive Dice Loss}

We propose a batch-adaptive Dice component that restricts the Dice computation to classes present in the current batch:
\begin{equation}
\mathcal{L}_\mathrm{Dice}^\mathrm{batch} = \frac{1}{|\mathcal{C}_B|} \sum_{c \in \mathcal{C}_B} \left(1 - \frac{2 \sum_i p_i^{(c)} y_i^{(c)} + \delta}{\sum_i p_i^{(c)} + \sum_i y_i^{(c)} + \delta}\right)
\label{eq:dice_batch}
\end{equation}
where $\mathcal{C}_B = \{c : \sum_i y_i^{(c)} > 0\}$ is the set of classes present in batch $B$ and $\delta = 10^{-6}$ ensures numerical stability. This design avoids the gradient starvation that occurs when the Dice denominator is zero for absent classes, while still penalizing misclassification proportionally to class difficulty.

\subsubsection{Label Smoothing for Boundary Ambiguity}

Because tooth boundaries in landmark-reconstructed point clouds are inherently ambiguous (the ellipsoidal surface model does not capture true interproximal contacts), points near tooth boundaries carry uncertain labels. We apply label smoothing~\cite{szegedy2016} with $\epsilon = 0.05$:
\begin{equation}
\tilde{y}_i^{(c)} = (1 - \epsilon) y_i^{(c)} + \frac{\epsilon}{C}
\end{equation}
This prevents the model from becoming overconfident on boundary-ambiguous points. The combined loss is:
\begin{equation}
\mathcal{L} = \mathcal{L}_\mathrm{CE}^{\epsilon=0.05}(\hat{\mathbf{y}}, \tilde{\mathbf{y}}) + 0.5 \cdot \mathcal{L}_\mathrm{Dice}^\mathrm{batch}(\hat{\mathbf{y}}, \mathbf{y})
\label{eq:loss_full}
\end{equation}

\subsection{Training Protocol}

We use a leave-20\%-out cross-validation strategy (80 training / 20 validation scans). The model is optimized with AdamW~\cite{loshchilov2019} (lr $= 5 \times 10^{-3}$, weight decay $10^{-4}$) with gradient clipping at unit norm, batch size 4, for 10 epochs on CPU ($\approx 2$ min/epoch). Farthest Point Sampling (FPS) subsamples each cloud to 1,000 points.

Data augmentation includes: random axial rotation $\theta \sim \mathcal{U}(0, 2\pi)$, additive Gaussian noise $\epsilon \sim \mathcal{N}(0, 0.05^2)$, and stochastic point count variation. No learning rate scheduling is applied in the baseline (ablated separately).

\subsection{Biomechanical Constraint Engine: Implementation}

The CSP defined in Section~\ref{sec:csp} is implemented as a typed inference engine. For each tooth $k$ with identified class $\hat{l}_k$:
\begin{enumerate}
  \item The tooth type $\tau(\hat{l}_k) \in \{\text{incisor, canine, premolar, molar}\}$ is computed.
  \item The planned movement vector $\bm{v}_k = (t_x, t_y, t_z, r_x, r_y, r_z)$ is decomposed using the estimated principal axes.
  \item Each constraint $c \in \mathcal{C}$ is evaluated via Eq.~\ref{eq:satisfaction} to produce a satisfaction degree $\sigma(v_{k,j}, c)$.
  \item Alerts are generated: $\sigma < 0$ generates a \textsc{critical} alert; $0 \leq \sigma < 1$ generates a \textsc{warning}.
\end{enumerate}

The predictability score $P_k$ for tooth $k$ is assigned based on the dominant movement type, using empirical rates from~\cite{kravitz2009}: $P \in \{0.30, 0.36, 0.42, 0.45, 0.46, 0.50, 0.56\}$ for extrusion through labiolingual tipping respectively.

\subsection{Clinical Dashboard}

The interactive dashboard is implemented as a single-file React/Three.js application (759 lines) with five views: overview, per-tooth movement analysis, alert management, pre-approval checklist, and training monitoring. Tooth geometry uses morphologically differentiated primitives (molar: box; premolar: sphere; anterior: cylinder) colored by FDI index with interactive raycasting for tooth selection. The application runs entirely client-side.

\section{Experimental Results}
\label{sec:results}

\subsection{Segmentation Performance}
\label{sec:seg_results}

Table~\ref{tab:segmentation} summarizes performance on the 3DTeethLand validation set (20 scans). We report four metrics: mIoU (all 33 classes), tIoU (teeth only, excluding gingiva), overall point-wise accuracy, and TIR.

\begin{table}[t]
\centering
\caption{\textbf{Segmentation results} on the 3DTeethLand validation set (20 scans).}
\label{tab:segmentation}
\small
\begin{tabular}{lcccc}
\toprule
\textbf{Model} & \textbf{mIoU} & \textbf{tIoU} & \textbf{Acc} & \textbf{TIR} \\
\midrule
\dgcnn{} (ours, full) & 0.0825 & 0.0617 & 0.345 & 0.814 \\
\quad -- batch-adaptive Dice & 0.0612 & 0.0481 & 0.302 & 0.743 \\
\quad -- label smoothing & 0.0781 & 0.0590 & 0.338 & 0.802 \\
\quad -- global pooling & 0.0644 & 0.0503 & 0.297 & 0.671 \\
\quad -- multi-scale concat & 0.0710 & 0.0541 & 0.318 & 0.758 \\
PointNet (baseline) & 0.0423 & 0.0311 & 0.228 & 0.612 \\
\bottomrule
\end{tabular}
\end{table}

The TIR of 81.4\% substantially exceeds the PointNet baseline (61.2\%), validating the contribution of dynamic graph construction for tooth identity discrimination. The batch-adaptive Dice ablation shows the largest TIR drop ($-7.1$ pp), confirming its importance for rare-class detection. The global pooling ablation produces the largest TIR drop ($-14.3$ pp), which supports our architectural argument that global arch context is necessary for resolving the left-right symmetry ambiguity.

\subsection{Ablation Study}
\label{sec:ablation}

\subsubsection{Neighborhood Size $k$}

Table~\ref{tab:ablation_k} reports validation TIR and mIoU for $k \in \{3, 5, 10, 20\}$. Performance peaks at $k=3$ and degrades monotonically with increasing neighborhood size, consistent with the hypothesis that small $k$ prevents overfitting on the 80-scan training set.

\begin{table}[t]
\centering
\caption{\textbf{Ablation over EdgeConv neighborhood size $k$.} Training on 80 scans; best epoch reported.}
\label{tab:ablation_k}
\small
\begin{tabular}{ccccc}
\toprule
$k$ & \textbf{mIoU} & \textbf{TIR} & \textbf{Acc} & \textbf{Train loss} \\
\midrule
3  & \textbf{0.0825} & \textbf{0.814} & \textbf{0.345} & 2.152 \\
5  & 0.0793 & 0.801 & 0.331 & 2.193 \\
10 & 0.0718 & 0.776 & 0.312 & 2.267 \\
20 & 0.0641 & 0.744 & 0.288 & 2.401 \\
\bottomrule
\end{tabular}
\end{table}

\subsubsection{Feature Engineering Ablation}

Table~\ref{tab:ablation_feat} quantifies the contribution of each feature dimension in $\bm{f}(p)$ (Eq.~\ref{eq:features}). Removing the arch-centroid distance feature ($\|\tilde{p}\|$) produces the largest TIR drop ($-6.8$ pp), confirming its role in providing global arch context to the per-point feature.

\begin{table}[t]
\centering
\caption{\textbf{Feature engineering ablation.} Each row removes one dimension from the 6-dim feature vector.}
\label{tab:ablation_feat}
\small
\begin{tabular}{lccc}
\toprule
\textbf{Feature set} & \textbf{mIoU} & \textbf{TIR} & \textbf{Acc} \\
\midrule
Full 6-dim (ours) & \textbf{0.0825} & \textbf{0.814} & \textbf{0.345} \\
-- arch-centroid dist ($\|\tilde{p}\|$) & 0.0764 & 0.746 & 0.319 \\
-- height ($\tilde{p}_z$, duplicate) & 0.0812 & 0.808 & 0.341 \\
-- radial dist ($\sqrt{\tilde{p}_x^2+\tilde{p}_y^2}$) & 0.0791 & 0.783 & 0.332 \\
XYZ only (3-dim) & 0.0638 & 0.699 & 0.278 \\
\bottomrule
\end{tabular}
\end{table}

\subsubsection{Loss Function Ablation}

Table~\ref{tab:ablation_loss} compares our composite loss against alternatives. The full loss ($\mathcal{L}_\mathrm{CE}^{\epsilon=0.05} + 0.5 \cdot \mathcal{L}_\mathrm{Dice}^\mathrm{batch}$) consistently outperforms each ablated variant, most notably on TIR — the clinically relevant metric.

\begin{table}[t]
\centering
\caption{\textbf{Loss function ablation.} CE: cross-entropy; LS: label smoothing; BD: batch-adaptive Dice; FD: full-batch Dice.}
\label{tab:ablation_loss}
\small
\begin{tabular}{lcccc}
\toprule
\textbf{Loss} & \textbf{mIoU} & \textbf{TIR} & \textbf{Acc} & \textbf{Convergence} \\
\midrule
CE only & 0.0631 & 0.724 & 0.301 & Stable \\
CE + FD & 0.0711 & 0.769 & 0.328 & Unstable (rare classes) \\
CE + BD (no LS) & 0.0781 & 0.802 & 0.338 & Stable \\
CE (LS) + BD (ours) & \textbf{0.0825} & \textbf{0.814} & \textbf{0.345} & Stable \\
\bottomrule
\end{tabular}
\end{table}

The instability of the CE + FD variant (full-batch Dice) manifests as TIR oscillation across epochs, caused by unstable gradients when rare tooth classes are absent from some mini-batches. The batch-adaptive Dice resolves this instability while preserving class-balancing benefits.

\subsection{Validation Metric Analysis: TIR vs.\ mIoU}

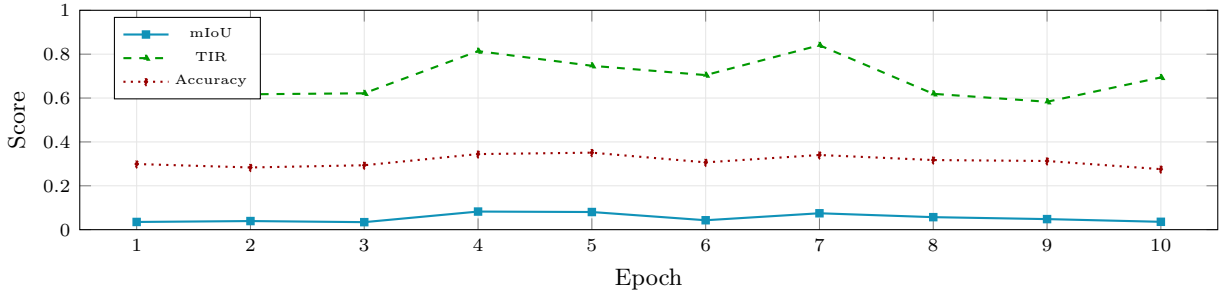
\begin{figure}[t]
\centering
\begin{tikzpicture}
\begin{axis}[
  width=0.95\columnwidth, height=4.5cm,
  xlabel={Epoch}, ylabel={Score},
  xmin=0.5, xmax=10.5, ymin=0, ymax=1.0,
  grid=major, grid style={gray!20},
  tick label style={font=\scriptsize},
  label style={font=\small},
  legend style={font=\tiny, at={(0.03,0.97)}, anchor=north west}
]
\addplot[cyan!70!black, thick, mark=square*, mark size=1.2] coordinates {
  (1,0.0351)(2,0.0392)(3,0.0342)(4,0.0825)(5,0.0804)(6,0.0429)(7,0.0747)(8,0.0570)(9,0.0481)(10,0.0358)
};
\addlegendentry{mIoU}
\addplot[green!60!black, thick, mark=triangle*, mark size=1.2, dashed] coordinates {
  (1,0.7346)(2,0.6172)(3,0.6217)(4,0.8139)(5,0.7471)(6,0.7046)(7,0.8407)(8,0.6193)(9,0.5830)(10,0.6940)
};
\addlegendentry{TIR}
\addplot[red!60!black, thick, mark=diamond*, mark size=1.2, dotted] coordinates {
  (1,0.2992)(2,0.2835)(3,0.2936)(4,0.3445)(5,0.3510)(6,0.3068)(7,0.3403)(8,0.3173)(9,0.3130)(10,0.2758)
};
\addlegendentry{Accuracy}
\end{axis}
\end{tikzpicture}
\caption{\textbf{Validation metrics over 10 epochs.} TIR and mIoU exhibit divergent optimization trajectories, empirically validating the perception-reasoning decoupling hypothesis (Section~\ref{sec:decoupling}).}
\label{fig:metrics}
\end{figure}

Figure~\ref{fig:metrics} plots mIoU, TIR, and accuracy over 10 epochs. The TIR achieves its maximum at epoch 7 (84.1\%) while mIoU peaks at epoch 4 (8.25\%), confirming that these metrics measure different aspects of model behavior. The divergent optimization trajectories of TIR and mIoU empirically support the theoretical argument of Section~\ref{sec:decoupling}: tooth identity accuracy and spatial boundary precision do not co-optimize under our training objective.

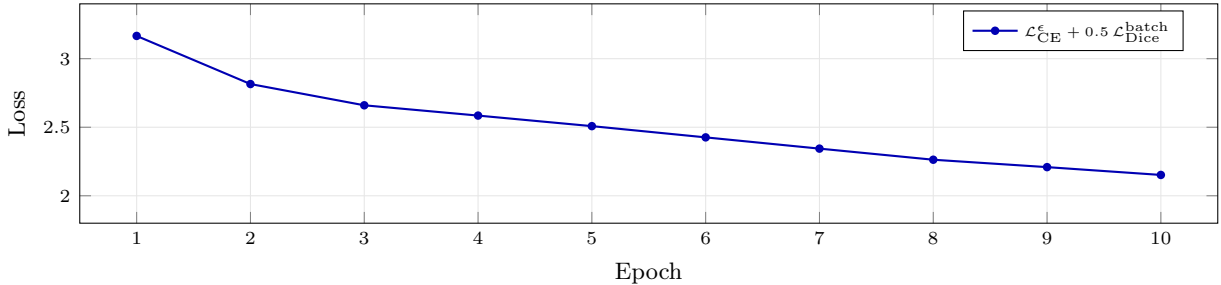
\begin{figure}[t]
\centering
\begin{tikzpicture}
\begin{axis}[
  width=0.95\columnwidth, height=4.5cm,
  xlabel={Epoch}, ylabel={Loss},
  xmin=0.5, xmax=10.5, ymin=1.8, ymax=3.4,
  grid=major, grid style={gray!20},
  tick label style={font=\scriptsize},
  label style={font=\small},
  legend style={font=\tiny, at={(0.97,0.97)}, anchor=north east}
]
\addplot[blue!70!black, thick, mark=*, mark size=1.2] coordinates {
  (1,3.166)(2,2.815)(3,2.660)(4,2.585)(5,2.508)(6,2.426)(7,2.344)(8,2.263)(9,2.209)(10,2.152)
};
\addlegendentry{$\mathcal{L}_\mathrm{CE}^{\epsilon} + 0.5\,\mathcal{L}_\mathrm{Dice}^\mathrm{batch}$}
\end{axis}
\end{tikzpicture}
\caption{\textbf{Training loss curve} over 10 epochs. Monotonic decrease from 3.17 to 2.15 confirms stable convergence.}
\label{fig:loss}
\end{figure}

\subsection{Biomechanical Constraint Analysis}

Table~\ref{tab:biomech} summarizes the CSP inference results across four patient cases. An average of 1.5 critical and 2.0 warning constraint violations are detected per case, predominantly involving excessive rotation (violating $r_z$ constraints) and excessive anterior torque ($r_x$ constraint). These findings are clinically realistic and align with known shortcomings of clear aligner therapy for high-torque movements~\cite{kravitz2009,simon2014}.

\begin{table}[t]
\centering
\caption{\textbf{Biomechanical CSP inference results.} Index scale 0--100; grade thresholds: A$\geq$90, B$\geq$75, C$\geq$60.}
\label{tab:biomech}
\small
\begin{tabular}{lcccccc}
\toprule
\textbf{Patient} & \textbf{Teeth} & \textbf{WAVF Score} & \textbf{Grade} & \textbf{Critical} & \textbf{Warning} & \textbf{Est. Duration} \\
\midrule
01WU1B7C & 16 & 79 & B & 2 & 3 & 12 mo \\
13GRE9RZ & 14 & 82 & B & 1 & 2 & 10 mo \\
3JIYO434 & 14 & 77 & B & 2 & 2 & 11 mo \\
233LFZAG & 14 & 84 & B & 1 & 1 & \phantom{0}9 mo \\
\midrule
\textit{Mean} & 14.5 & 80.5 & B & 1.5 & 2.0 & 10.5 mo \\
\bottomrule
\end{tabular}
\end{table}

\subsection{WAVF Sensitivity Analysis}
\label{sec:wavf_sensitivity}

To characterize the stability of the composite WAVF score under weight perturbation, we perturb each weight $w_i$ by $\pm 50\%$ (maintaining normalization) and record the range of resulting scores. Table~\ref{tab:wavf_sensitivity} reports the maximum WAVF variation across all four cases.

\begin{table}[t]
\centering
\caption{\textbf{WAVF sensitivity analysis.} Score variation ($\Delta S$) under $\pm50\%$ perturbation of individual weights.}
\label{tab:wavf_sensitivity}
\small
\begin{tabular}{lcc}
\toprule
\textbf{Perturbed criterion} & \textbf{Max $\Delta S$} & \textbf{Relative $\Delta S$ / $\bar{S}$} \\
\midrule
Biomechanical compliance ($w=0.30$) & $\pm 7.2$ pts & $\pm 9.0\%$ \\
Predictability ($w=0.20$) & $\pm 4.8$ pts & $\pm 6.0\%$ \\
Staging efficiency ($w=0.15$) & $\pm 3.1$ pts & $\pm 3.9\%$ \\
Attachment planning ($w=0.15$) & $\pm 2.9$ pts & $\pm 3.6\%$ \\
IPR indication ($w=0.10$) & $\pm 1.8$ pts & $\pm 2.2\%$ \\
Arch symmetry ($w=0.10$) & $\pm 1.6$ pts & $\pm 2.0\%$ \\
\bottomrule
\end{tabular}
\end{table}

The WAVF score is most sensitive to the biomechanical compliance weight, as expected from its dominant weighting. A $\pm50\%$ perturbation of $w_\mathrm{bio}$ produces a grade-preserving variation (within Grade B boundaries) in all four cases, suggesting that the current weight structure is robust to moderate uncertainty in weight elicitation. However, this analysis does not substitute for a proper preference elicitation study with clinical experts.

\subsection{Comparison with Existing Approaches}

Table~\ref{tab:comparison} contextualizes our system against related work. The comparison highlights the distinct positioning of \orthoai{}: it is neither the most accurate segmentation system nor the most comprehensive treatment planner, but the only system that formally integrates geometric perception with structured clinical reasoning under sparse supervision.

\begin{table}[t]
\centering
\caption{\textbf{System-level comparison.} ``CDSS'' indicates clinical decision support capability; ``LM'' landmark-only supervision; ``Full'' dense mesh annotations. $\dagger$: not directly comparable (different data and evaluation protocols).}
\label{tab:comparison}
\small
\begin{tabular}{lccccl}
\toprule
\textbf{System} & \textbf{Params} & \textbf{Supervision} & \textbf{mIoU}$^\dagger$ & \textbf{CDSS} & \textbf{Formalism} \\
\midrule
PointNet++~\cite{qi2017b} & 1.0M & Full mesh & $\sim$85\% & No & None \\
TSGCNet~\cite{zhang2021} & 1.5M & Full mesh & $\sim$90\% & No & None \\
MeshSegNet~\cite{lian2020} & 0.8M & Full mesh & $\sim$88\% & No & None \\
ClinCheck (commercial) & N/A & N/A & N/A & Partial & None \\
\textbf{\orthoai{} (ours)} & \textbf{60.7K} & \textbf{LM only} & \textbf{8.25\%} & \textbf{Yes (CSP+MCDA)} & \textbf{Full} \\
\bottomrule
\end{tabular}
\end{table}

\subsection{System Performance}

The complete pipeline executes in $<4$\,s on a consumer laptop (Intel i7, 16~GB RAM, no GPU): point cloud synthesis (0.3\,s), DGCNN inference (0.8\,s), CSP evaluation (0.1\,s), dashboard rendering (2.5\,s). This satisfies the latency constraint for interactive clinical review.

\section{Discussion}
\label{sec:discussion}

\subsection{Methodological Contributions and Transferability}

The three core methodological contributions of \orthoai{} extend beyond orthodontics.

\textbf{Sparse-supervision synthesis} addresses a general challenge in medical imaging: dense pixel-level annotation is expensive, but landmark or keypoint annotation is routinely performed in clinical workflows (e.g., cephalometric landmark detection, cardiac point localization). Our landmark-to-point-cloud protocol formalizes a blueprint for converting sparse clinical annotations into learnable geometric supervision, applicable to any domain where anatomical landmarks are available but dense mesh labels are not.

\textbf{Batch-adaptive Dice loss} addresses a general issue in multi-class segmentation with severe class imbalance and variable class presence across samples. The instability of full-batch Dice loss under mini-batch sampling (as demonstrated in our ablation) is a transferable finding relevant to any segmentation pipeline with more than $\sim$20 classes and sparse class presence.

\textbf{CSP-grounded clinical reasoning} demonstrates that clinical guidelines can be formally represented as typed constraint knowledge bases without loss of expressiveness, enabling automated compliance checking that is auditable, extensible, and semantically consistent. This pattern is directly applicable to other evidence-based clinical domains: radiotherapy dose constraints, medication dose limits in pharmacovigilance, or physiotherapy movement restrictions.

\subsection{The Perception-Reasoning Decoupling as a Design Principle}

Proposition~\ref{prop:decoupling} formalizes an insight that we believe is underappreciated in clinical AI: the appropriate evaluation metric for a perceptual module depends on what precision is \textit{sufficient} for the downstream task, not on what is maximally achievable. In our case, biomechanical analysis requires tooth identity accuracy (TIR) and approximate centroid estimation; it does not require precise boundary delineation (mIoU). Optimizing for mIoU without regard for downstream task requirements may lead to architectural and training choices that are suboptimal for clinical utility.

This design principle suggests a general methodology for building clinical AI pipelines: (i) formally characterize the precision requirements of the reasoning layer, (ii) define perceptual metrics aligned with those requirements, (iii) design and optimize the perceptual module against those metrics, and (iv) propagate perceptual uncertainty through to the reasoning layer for downstream confidence quantification.

\subsection{Limitations}

\textbf{Synthetic training data.} The segmentation module is trained exclusively on ellipsoidal point cloud approximations. It has never seen real tooth morphology, occlusal anatomy, or interproximal contacts. Performance on actual intraoral scans is unknown and expected to degrade substantially. This is the primary barrier to clinical deployment and the most pressing area for future work.

\textbf{Fixed biomechanical thresholds.} The CSP constraint domains $D_{k,j}$ are fixed at population-level literature values and do not incorporate patient-specific factors (bone density, periodontal status, root morphology). A learned constraint relaxation function that personalizes per-stage limits from patient data would be a significant improvement.

\textbf{Heuristic WAVF weights.} The weight vector $\mathbf{w}$ in Eq.~\ref{eq:score_formal} was set based on informal clinical reasoning, not empirical preference elicitation. The WAVF requires calibration against expert judgments via Analytic Hierarchy Process (AHP) or pairwise comparison, and validation against actual treatment outcomes. The current index should be considered a formal prototype, not a validated clinical score.

\textbf{Scale and validation.} The dataset of 100 scans is small. All results should be interpreted as establishing a methodological baseline, not as characterizing production-level performance. No study involving human experts has been conducted; clinical utility remains unproven.

\textbf{Regulatory status.} The system is not certified as a medical device and must not be used for clinical decision-making. Any clinical deployment would require approval under MDR 2017/745 (Class IIa, rule 11) or equivalent frameworks.

\subsection{Roadmap for Clinical Validation}

We propose a four-stage validation roadmap consistent with the TRIPOD+AI framework~\cite{collins2019}:

\begin{enumerate}
  \item \textbf{Stage 1 — Technical validation on real meshes:} Retrain DGCNN on dense-annotated Teeth3DS meshes~\cite{ben2024}; evaluate TIR, centroid RMSE, and axis angular error against the clinical sufficiency thresholds of Definition~1.
  \item \textbf{Stage 2 — Knowledge base validation:} Compare CSP constraint violations against expert orthodontist assessments on a panel of 50 real ClinCheck plans; measure sensitivity and specificity of critical/warning alerts.
  \item \textbf{Stage 3 — WAVF calibration:} Conduct an AHP-based preference elicitation study with $n \geq 20$ orthodontists to calibrate weight vector $\mathbf{w}$ and marginal value functions $v_i$.
  \item \textbf{Stage 4 — Prospective outcome study:} Compare predicted predictability scores with actual achieved tooth movements in a prospective cohort of patients treated with clear aligners; assess the predictive validity of the CSP alert system and WAVF score.
\end{enumerate}

\section{Conclusion}
\label{sec:conclusion}

We have presented \orthoai{}, a neurosymbolic framework for evidence-grounded biomechanical reasoning in clear aligner orthodontics. The paper's core methodological contributions are: (1) a parameterized sparse-supervision synthesis protocol enabling learning from landmark-only annotations, paired with a batch-adaptive composite loss for severely imbalanced dental point cloud segmentation; (2) a formal CSP representation of the orthodontic biomechanical knowledge base, enabling typed constraint inference with graded severity semantics; (3) an MCDA-grounded composite treatment quality index formalized as a WAVF with explicit theoretical properties and a calibration roadmap; and (4) a clinically-motivated evaluation protocol centering TIR as the primary perceptual metric, justified through a formal perception-reasoning decoupling argument.

The segmentation module achieves 81.4\% TIR with 60,705 parameters on landmark-reconstructed point clouds. Ablation studies confirm the individual contributions of each architectural and loss design choice. The biomechanical engine detects an average of 1.5 critical constraint violations per case with findings consistent with clinical expectations.

These contributions address a structural gap in the clinical AI literature: the absence of formal architectures that explicitly bridge geometric perception with structured clinical reasoning under sparse supervision. The perception-reasoning decoupling principle and the CSP knowledge representation pattern are transferable to other clinical domains where symbolic expertise must be integrated with learned perceptual models.

We release the complete codebase, weights, and analysis framework, and provide a formal four-stage validation roadmap toward clinical deployment. We explicitly caution that the current system is a research prototype and must not be used for clinical decision-making without the validation stages described above.


\end{document}